\documentclass[10pt,logo,copyright]{giga-report}
\usepackage{iftex}
\ifPDFTeX
    \usepackage[utf8]{inputenc} %
    \usepackage[T1]{fontenc}    %
\fi

\usepackage{parskip}        %
\usepackage{url}            %
\usepackage{booktabs}       %
\usepackage{amsfonts}       %
\usepackage{nicefrac}       %
\usepackage{microtype}      %
\usepackage[dvipsnames]{xcolor} %
\usepackage{graphicx}
\usepackage{animate}        %
\usepackage{subcaption}
\usepackage{tabularx}
\usepackage{makecell}
\usepackage{adjustbox}
\usepackage{setspace}
\usepackage{todonotes}
\usepackage{colortbl}
\usepackage{wrapfig}

\definecolor{rankbest}{RGB}{144,238,144}
\definecolor{rankmid}{RGB}{255,255,180}
\definecolor{rankworst}{RGB}{248,150,150}
\definecolor{deltadarkpos}{RGB}{50,180,50}
\definecolor{deltalightpos}{RGB}{180,235,180}
\definecolor{deltayellow}{RGB}{255,255,200}
\definecolor{deltalightneg}{RGB}{250,180,180}
\definecolor{deltadarkneg}{RGB}{220,80,80}

\newcolumntype{M}[1]{>{\centering\arraybackslash}m{#1}}
\usepackage{float}
\usepackage{tikz}
\usetikzlibrary{positioning,shapes,arrows,arrows.meta,fit,tikzmark}
\usepackage{amsmath,amsfonts,bm, bbm,leftindex}
\usepackage{multirow}
\usepackage{comment}
\usepackage{lipsum}
\usepackage[para]{threeparttable}

\def\eqref#1{equation~\ref{#1}}

\def\1{\bm{1}}

\DeclareMathAlphabet{\mathsfit}{\encodingdefault}{\sfdefault}{m}{sl}
\SetMathAlphabet{\mathsfit}{bold}{\encodingdefault}{\sfdefault}{bx}{n}

\makeatletter
\let\save@mathaccent\mathaccent
\newcommand*\if@single[3]{%
  \setbox0\hbox{${\mathaccent"0362{#1}}^H$}%
  \setbox2\hbox{${\mathaccent"0362{\kern0pt#1}}^H$}%
  \ifdim\ht0=\ht2 #3\else #2\fi
  }
\newcommand*\rel@kern[1]{\kern#1\dimexpr\macc@kerna}
\newcommand*\widebar[1]{\@ifnextchar^{{\wide@bar{#1}{0}}}{\wide@bar{#1}{1}}}
\newcommand*\wide@bar[2]{\if@single{#1}{\wide@bar@{#1}{#2}{1}}{\wide@bar@{#1}{#2}{2}}}
\newcommand*\wide@bar@[3]{%
  \begingroup
  \def\mathaccent##1##2{%
    \let\mathaccent\save@mathaccent
    \if#32 \let\macc@nucleus\first@char \fi
    \setbox\z@\hbox{$\macc@style{\macc@nucleus}_{}$}%
    \setbox\tw@\hbox{$\macc@style{\macc@nucleus}{}_{}$}%
    \dimen@\wd\tw@
    \advance\dimen@-\wd\z@
    \divide\dimen@ 3
    \@tempdima\wd\tw@
    \advance\@tempdima-\scriptspace
    \divide\@tempdima 10
    \advance\dimen@-\@tempdima
    \ifdim\dimen@>\z@ \dimen@0pt\fi
    \rel@kern{0.6}\kern-\dimen@
    \if#31
      \overline{\rel@kern{-0.6}\kern\dimen@\macc@nucleus\rel@kern{0.4}\kern\dimen@}%
      \advance\dimen@0.4\dimexpr\macc@kerna
      \let\final@kern#2%
      \ifdim\dimen@<\z@ \let\final@kern1\fi
      \if\final@kern1 \kern-\dimen@\fi
    \else
      \overline{\rel@kern{-0.6}\kern\dimen@#1}%
    \fi
  }%
  \macc@depth\@ne
  \let\math@bgroup\@empty \let\math@egroup\macc@set@skewchar
  \mathsurround\z@ \frozen@everymath{\mathgroup\macc@group\relax}%
  \macc@set@skewchar\relax
  \let\mathaccentV\macc@nested@a
  \if#31
    \macc@nested@a\relax111{#1}%
  \else
    \def\gobble@till@marker##1\endmarker{}%
    \futurelet\first@char\gobble@till@marker#1\endmarker
    \ifcat\noexpand\first@char A\else
      \def\first@char{}%
    \fi
    \macc@nested@a\relax111{\first@char}%
  \fi
  \endgroup
}
\makeatother

\definecolor{darkred}{rgb}{0.7, 0.0, 0.0}

\usepackage{amsmath} 
\usepackage{marvosym}

\usepackage[table]{xcolor}

\usepackage{pifont}
\usepackage{graphicx}

\usepackage[nameinlink]{cleveref}
\crefname{equation}{Eq.}{Eqs.}
\crefname{figure}{Fig.}{Figs.}
\crefname{section}{Sec.}{Sec.}
\crefname{appendix}{App.}{App.}
\crefname{table}{Tab.}{Tabs.}
\crefname{algorithm}{Algo}{Algo}
\crefname{thm}{Thm}{Thm}
\Crefname{thm}{Thm}{Thm}
\crefname{prop}{Prop}{Prop}

\newcommand{\crefnames}[3]{%
  \@for\next:=#1\do{%
    \expandafter\crefname\expandafter{\next}{#2}{#3}%
  }%
}

\title{GigaWorld-Policy-0.5: A Faster and Stronger WAM Empowered by AutoResearch}

\author{
\vspace{-0.1in}
\centerline{GigaAI}
\centerline{Tsinghua University}
\centerline{{Project Page: \href{https://open-gigaai.github.io/giga-world-policy}{https://open-gigaai.github.io/giga-world-policy/}}}
\footnotesize
\textbf{Alphabetical Order}:
\normalfont
  Angen Ye,
  Angyuan Ma,
  Boyuan Wang,
  Chaojun Ni,
  Fangzheng Ye,
  Guan Huang, 
  Guo Li,
  Guosheng Zhao,\newline
  Haodong Yan,
  Hengtao Li,
  Jiwen Lu,
  Kai Wang,
  Mingming Yu,
  Qitang Hu,
  Qiuping Deng,
  Songling Liu,
  Xiaoyu Tian,
  Xiaofeng Wang, \newline
  Xinyu Zhou,
  Xiuwei Xu,
  Xinze Chen,
  Yang Wang,
  Yejun Zeng,
  Yifan Chang,
  Yun Ye,
  Zhenyu Wu,
  Zhanqian Wu,
  Zheng Zhu
\vspace{-1.5em}
}
\usepackage[numbers,sort&compress]{natbib}
\setcitestyle{numbers}
\begin{document}
\maketitle

\begin{abstract}
World Action Models (WAMs) improve robot policy learning by jointly modeling actions and future visual observations, using future scene evolution as dense supervision for physically grounded action generation. However, a common design in existing WAMs is to explicitly generate future videos at inference time, incurring substantial computational overhead and hindering real-time closed-loop deployment. GigaWorld-Policy addresses this issue with an action-centered formulation, where future visual dynamics are used during training while action-only decoding is used at inference time. Building upon this framework, we present \textit{GigaWorld-Policy-0.5}, an enhanced action-centered WAM designed for more efficient robot control. During pretraining, \textit{GigaWorld-Policy-0.5} adopts a mixed Action-Conditioned World Modeling (AC-WM) and WAM training strategy. This strengthens the coupling between visual dynamics and robot actions and improves the transferability of action representations for downstream policy learning. For efficient inference, \textit{GigaWorld-Policy-0.5} introduces a Mixture-of-Transformers architecture that separates visual dynamics modeling and action generation into specialized experts, reducing active computation during action-only inference and achieving 85 ms inference latency on a local RTX 4090 setup. In addition, we employ an agent-based AutoResearch pipeline to systematically search training configurations, enabling more efficient identification of optimal experimental setups while reducing the time and manual intervention required for hyperparameter tuning. Experiments and ablations show that \textit{GigaWorld-Policy-0.5} preserves the training benefits of future visual dynamics while improving inference efficiency for robot control.
\end{abstract}

\abscontent
\section{Introduction}
\label{sec:intro}

\begin{wrapfigure}{R}{0.48\textwidth}
    \centering
    \vspace{-6em}
    \captionsetup{type=figure, justification=justified, singlelinecheck=false}
    \includegraphics[width=\linewidth]{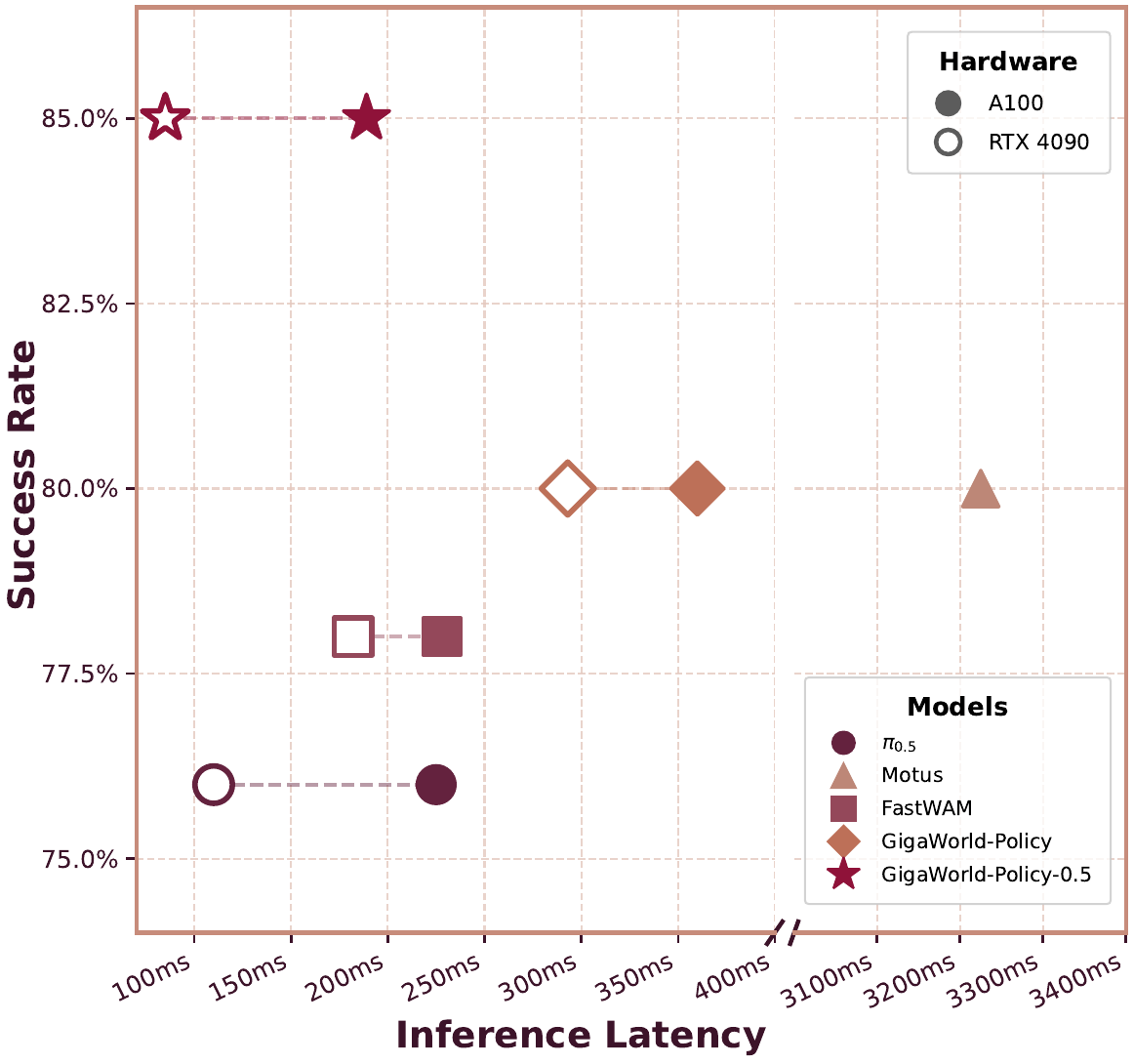}
    \vspace{-2em}
    \caption{Comparison of \textit{GigaWorld-Policy-0.5} with baselines on inference frequency and success rate across hardware platforms in real-world settings.}
    \vspace{-3em}
    \label{fig:introduction_overview}
\end{wrapfigure}

World models have recently made significant progress in learning predictive representations of the physical world from large-scale visual data~\citep{wan,cosmos,cosmos3,drivedreamer,drivedreamer2,gigaworld0,worldmodel, liu2025timestep, zhu2024sora,wang2024worlddreamer}. By modeling how scenes evolve over time, these models capture rich spatiotemporal priors about object motion, interaction dynamics, and long-horizon state transitions. Such predictive capability is naturally attractive for robotic control, where a policy must not only recognize the current observation, but also anticipate how its actions may change the environment.

Motivated by this progress, recent Vision-Language-Action (VLA) models have started to leverage world models to improve policy learning~\citep{pi0.7,gigabrain0,gigabrain0.5m,wog,beingh0.7,worldvla}. Compared with prior VLA methods~\citep{pi0,pi0.5,pi0.6,gigabrain0,gigabrain0.5m,swiftvla,gr00t,walloss,lingbotvla, ye2025vla} that primarily rely on sparse action supervision from robot demonstrations, world models provide complementary predictive information about how the scene may evolve. Incorporating such predictive information into VLA policies provides richer temporal context for action prediction and helps ground policy learning in future scene evolution. This line of work suggests that predictive world knowledge is an important ingredient for building more capable robot policies.

Beyond incorporating world model predictions into VLA policies, another line of work explores a more unified formulation through World Action Models (WAMs), which jointly model robot actions and future observations~\citep{motus,motubrain,lingbotva, ye2026halo,lv2026viva}. Instead of treating future prediction as an external input or auxiliary signal to a policy model, WAMs integrate action generation and future-dynamics modeling within the same framework. This formulation allows action representations to be learned together with their visual consequences: the model predicts not only what action should be executed, but also how the scene is expected to evolve under that action. As a result, future visual dynamics can serve as dense supervision for learning physically grounded actions.

While this unified formulation enables WAMs to acquire certain zero-shot generalization capabilities by coupling action generation with future scene evolution~\citep{dreamzero}, it also introduces a practical deployment challenge. In many existing WAM approaches, the same joint modeling process used for training is also invoked during inference, requiring explicit future-video generation, iterative denoising, or predictive rollout at deployment time~\citep{videovla,dreamzero}. Since video tokens are substantially more expensive than action tokens, this design introduces significant computational overhead and limits real-time closed-loop control. Moreover, errors in imagined future frames may accumulate over long horizons and weaken downstream action generation. Therefore, a key challenge is how to preserve the generalization benefit of action-dynamics coupling while avoiding costly future rollout during inference.

GigaWorld-Policy~\citep{gigaworldpolicy} addresses this challenge with an efficient action-centered World Action Model, following the broader direction of decoupling future prediction between training and inference~\citep{fastwam,gigaworldpolicy}. It leverages future visual dynamics to provide dense supervision during training, while allowing the policy to skip explicit future-video generation and directly decode actions at inference time. Through an action-centered causal token structure, action prediction is separated from future-video prediction, retaining the benefit of WAM training while substantially reducing inference latency for closed-loop deployment. As a result, GigaWorld-Policy demonstrates that WAMs can achieve low-latency action generation comparable to efficient VLA policies while retaining the stronger supervision and physical grounding provided by future visual dynamics.

Building upon GigaWorld-Policy, we present \textit{GigaWorld-Policy-0.5}. \textit{GigaWorld-Policy-0.5} inherits the core formulation of GigaWorld-Policy, where future visual dynamics are used to provide dense supervision during training, while inference can be performed through action-only decoding without explicit future-video generation. It also preserves the action-centered causal token structure: action tokens are predicted from the current observation, robot state, and language instruction, while future visual tokens are predicted conditioned on the action sequence. This masking strategy prevents future visual information from leaking into action prediction, and allows future-video prediction to remain optional during deployment, thereby retaining the low-latency control advantage of GigaWorld-Policy.

Compared with GigaWorld-Policy, \textit{GigaWorld-Policy-0.5} introduces several updates to the model architecture and training pipeline. First, it adopts a Mixture-of-Transformers (MoT) architecture that separates visual dynamics modeling and action generation into specialized experts while preserving the action-centered WAM design. Although MoT increases the total parameter count, its expert specialization enables an action-only inference pathway that avoids unnecessary visual-dynamics computation, achieving an inference latency of approximately 85 ms on an NVIDIA RTX 4090 and thereby improving deployment efficiency. Second, during pretraining, we adopt a mixed training strategy that combines Action-Conditioned World Modeling (AC-WM) with World Action Model training. This encourages the model to better capture the relationship between visual dynamics and robot actions, learning both general scene evolution and action-induced future changes, which in turn strengthens its action modeling capability. Finally, we employ an agent-based AutoResearch~\cite{karpathy2026autoresearch} pipeline to make the experimental search process more systematic and efficient. The pipeline automatically launches pilot runs, monitors validation metrics, compares candidate configurations, and promotes promising settings to longer training. Through this iterative process, AutoResearch helps derive a reliable training recipe for \textit{GigaWorld-Policy-0.5} with reduced manual intervention.
\section{Related Work}
\subsection{World Models as Data Engines for Robotics}
Recent advances in world models~\citep{li2025mimicdreamer,ni2025recondreamer,zhao2025drivedreamer4d,unidrivedreamer,drivedreamerpolicy,wang2026egovid} have improved robotic video generation and prediction~\citep{dong2025emma,liu2026robotransfer,swiftvla,wang2025embodiedreamer,Robodreamer} and have increasingly served as scalable data engines or learned simulators for robot learning~\citep{liu2026robotransfer, alhaija2025cosmos, gigaworld0, jiang2026wovr, wang2026reconphys}.
The central goal is to learn a generative or predictive model that captures the temporal evolution of embodied environments, enabling the synthesis or imagination of diverse visual experiences from limited real-world or simulated trajectories.

At the level of action-conditioned future prediction, Pandora~\citep{xiang2024pandora} demonstrates that free-form text actions can be used to control video world models, enabling interactive prediction beyond passive video generation. FreeAction~\citep{kim2025freeaction} further considers continuous robot actions and introduces action-scaled classifier-free guidance to control the intensity of generated motions. Building on such controllability, subsequent efforts have explored scalable generation and transfer of embodied visual data. RoboTransfer~\citep{liu2026robotransfer} and Cosmos-Transfer1~\citep{alhaija2025cosmos} focus on transferring visual experiences across domains, leveraging 3D or multimodal spatial conditions to improve geometric consistency and Sim2Real data generation. GigaWorld-0~\citep{gigaworld0} provides a high-fidelity video-and-3D data engine that synthesizes temporally coherent embodied trajectories with fine-grained control over appearance, scene layouts, viewpoints, and action semantics.  Qwen-RobotWorld~\citep{zhang2026qwen} further scales action-conditioned embodied video generation across multiple robotics domains, including manipulation, navigation, and driving. Going beyond video generation alone, Aether~\citep{zhu2025aether} unifies geometry-aware world modeling by jointly optimizing 4D dynamic reconstruction, action-conditioned video prediction, and goal-conditioned visual planning.

However, most existing efforts use world models primarily as external data engines or simulators~\citep{wang2025humandreamer, recondreamer++}, while largely overlooking how their predictive priors can be embedded into deployable robot policies.
This motivates recent work on world models for robot control, where future prediction and action generation are jointly considered.

\subsection{World Models for Robot Control}
World models have long been studied as predictive representations for decision making, and recent progress in large-scale video generation has renewed their relevance to robotic control.
Instead of mapping observations directly to actions, world-model-based policies exploit predictions of future scene evolution to provide temporal structure, dynamics priors, or intermediate plans.
Early video-based policy methods, such as UniPi~\citep{du2023learning}, formulate control as future video generation followed by action recovery, while generative robot models such as GR-2~\citep{wu2024unleashing} further combine large-scale video pretraining with action prediction for generalizable manipulation.

World Action Models~\citep{motus,lingbotva,videovla,gigaworldpolicy}, grounded in the video generation paradigm, aim to model robot actions and future visual dynamics within a unified framework.
By coupling action prediction with future visual modeling, WAMs provide dense temporal supervision and learned predictive priors beyond sparse demonstration actions.
VideoVLA~\citep{videovla} adapts pretrained video generation models into robotic manipulators, jointly predicting future visual outcomes and action sequences.
Motus~\citep{motus} introduces a unified latent action world model with a Mixture-of-Transformer architecture and UniDiffuser-style scheduling, enabling multiple modes including world modeling, inverse dynamics, policy learning, video generation, and joint video-action prediction. 
UWM~\citep{zhu2025unified} similarly couples video and action diffusion within a unified framework, showing that heterogeneous video and robot data can support scalable policy learning. 
Building on this direction, MotuBrain~\citep{motubrain} extends Motus with multiview modeling, cross-embodiment action representations, and deployment-oriented optimization for real-time control. 

Another branch investigates how much future prediction must be explicitly realized during policy inference. Mimic-video~\citep{pai2025mimic} uses an Internet-scale video backbone to produce latent video plans, which are decoded into actions through a flow-matching inverse-dynamics model. LingBot-VA~\citep{lingbotva} adopts autoregressive video-action world modeling with closed-loop rolling prediction to mitigate error accumulation during long-horizon execution. S-VAM~\citep{yan2026s} distills multi-step video foresight into efficient geometric and semantic representations, DiT4DiT~\citep{ma2026dit4dit} uses intermediate denoising features rather than decoded frames for action prediction, and LaWAM~\citep{chen2026lawam} exposes future dynamics through compact latent visual subgoals instead of pixel-level video. 

Complementary to these approaches, GigaWorld-Policy~\citep{gigaworldpolicy} and Fast-WAM~\citep{fastwam} explore more efficient ways to incorporate predictive world knowledge into action policies. Both methods use video-based predictive supervision during training while avoiding the need to explicitly generate future videos at test time. Together, these works indicate that the benefits of world modeling need not rely on computationally expensive pixel-level future prediction during policy execution.

Overall, prior work has explored a spectrum of designs that reduce the reliance on explicit pixel-level future rollout during policy inference. However, achieving the low latency required for closed-loop control on edge devices remains challenging. This motivates \textit{GigaWorld-Policy-0.5}, which retains an action-centered world action modeling formulation while improving inference efficiency through a lightweight action-only execution pathway.
\section{Method}

\begin{figure}[ht]
\centering
\captionsetup{type=figure, justification=justified, singlelinecheck=false}
\includegraphics[width=\textwidth]{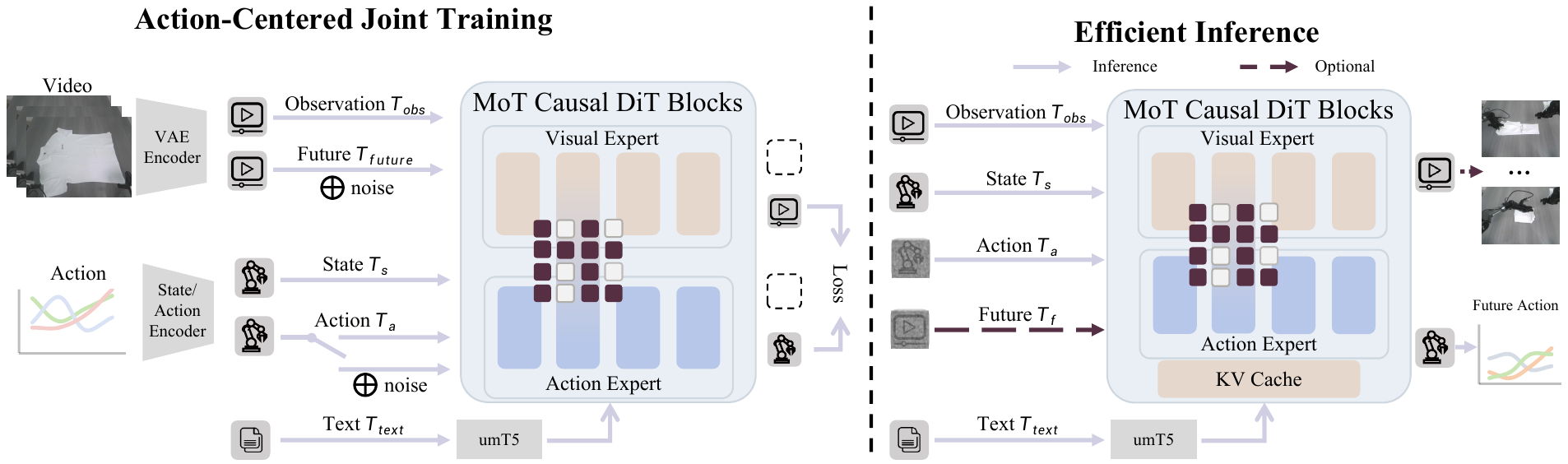}
\caption{
Overview of \textit{GigaWorld-Policy-0.5}, an MoT-based action-centered World Action Model. The model consists of a visual expert and an action expert: the visual expert specializes in processing video tokens, while the action expert focuses on action-token modeling. The two experts are connected through multi-modal self-attention, which follows the same causal masking strategy as GigaWorld-Policy~\citep{gigaworldpolicy} to preserve action-centered dependency modeling.
}
\label{fig:framework}
\end{figure}

\subsection{Overview of \textit{GigaWorld-Policy-0.5}}

The overall framework of \textit{GigaWorld-Policy-0.5} is illustrated in Fig.~\ref{fig:framework}. It follows the formulation of the action-centered world action model introduced in GigaWorld-Policy~\citep{gigaworldpolicy}. The policy prediction problem under this action-centered setting can be formulated as follows. Given the current multi-view observation $\{o_t^{\mathrm{left}}, o_t^{\mathrm{front}}, o_t^{\mathrm{right}}\}$, proprioceptive state $s_t$, and language instruction $l$, the model jointly predicts an action chunk $\hat{\mathbf{a}}_{t:t+p-1}$ of length $p$ and future visual observations $\hat{\mathbf{o}}_{t+\Delta:t+K\Delta}^{\mathrm{comp}}$:
\begin{equation}
    \left(
    \hat{\mathbf{a}}_{t:t+p-1},
    \hat{\mathbf{o}}_{t+\Delta:t+K\Delta}^{\mathrm{comp}}
    \right)
    \sim
    g_{\theta}
    \left(
    \cdot
    \mid
    o_t^{\mathrm{comp}}, s_t, l
    \right),
\end{equation}
where
\begin{equation}
    \hat{\mathbf{a}}_{t:t+p-1}
    =
    \left(\hat{a}_t, \hat{a}_{t+1}, \ldots, \hat{a}_{t+p-1}\right),
    \quad
    \hat{\mathbf{o}}_{t+\Delta:t+K\Delta}^{\mathrm{comp}}
    =
    \left(
    \hat{o}_{t+\Delta}^{\mathrm{comp}},
    \hat{o}_{t+2\Delta}^{\mathrm{comp}},
    \ldots,
    \hat{o}_{t+K\Delta}^{\mathrm{comp}}
    \right).
\end{equation}
Here, $\Delta$ denotes the temporal stride for future visual prediction, and $K=\lfloor p/\Delta \rfloor$ denotes the number of predicted future observations within the action horizon.

The composite observation $o^{\mathrm{comp}}$ is constructed by concatenating the left, front, and right camera views:
\begin{equation}
    o^{\mathrm{comp}}
    =
    \mathrm{Compose}
    \left(
        o^{\mathrm{left}},
        o^{\mathrm{front}},
        o^{\mathrm{right}}
    \right).
\end{equation}

Notably, \textit{GigaWorld-Policy-0.5} adopts a similar multi-view composition strategy as Motus~\citep{motus}, where the front view is placed on the top and the left and right views are concatenated in the bottom row to form a unified image input for visual encoding.

To better transfer world-model priors into policy learning, \textit{GigaWorld-Policy-0.5} adopts the same causal masking strategy as GigaWorld-Policy~\citep{gigaworldpolicy}. Specifically, future visual tokens are allowed to attend to action tokens, while action tokens are prevented from attending to future visual tokens. This design has two benefits. First, during training, the causal attention pattern implicitly conditions future visual prediction on the action tokens, encouraging the model to learn action representations that are consistent with the world-model prior and the expected scene evolution. Second, during inference, future visual prediction is optional; when low-latency control is desired, the model can skip the large number of future visual tokens and directly decode action tokens.

\textit{GigaWorld-Policy-0.5} is optimized with the flow matching framework~\citep{rectifiedflow, lipman2022flow}. 
For action tokens and future visual tokens, we sample modality-specific flow timesteps with different flow-shift factors. 
Given $r^{\mathrm{a}}, r^{\mathrm{v}} \sim \mathcal{U}(0,1)$ and flow-shift factors $\gamma_{\mathrm{a}}$ and $\gamma_{\mathrm{v}}$, the shifted timesteps are computed as
\begin{equation}
    \tau^{\mathrm{a}}
    =
    \frac{\gamma_{\mathrm{a}} r^{\mathrm{a}}}
    {1 + (\gamma_{\mathrm{a}} - 1) r^{\mathrm{a}}},
    \quad
    \tau^{\mathrm{v}}
    =
    \frac{\gamma_{\mathrm{v}} r^{\mathrm{v}}}
    {1 + (\gamma_{\mathrm{v}} - 1) r^{\mathrm{v}}}.
\end{equation}
We then form a joint timestep vector for the action and future visual modalities:
\begin{equation}
    \tau=
    \begin{bmatrix}
        \tau^{\mathrm{a}}\\
        \tau^{\mathrm{v}}
    \end{bmatrix}.
\end{equation}

Let $x_1^{\mathrm{a}}$ and $x_1^{\mathrm{v}}$ denote the clean action tokens and future visual tokens, respectively, and let $x_0^{\mathrm{a}}$ and $x_0^{\mathrm{v}}$ denote their corresponding Gaussian noise. 
The noisy joint token is constructed by linearly interpolating between noise and data for each modality:
\begin{equation}
    x_{\tau}
    =
    \tau
    \begin{bmatrix}
         x_1^{\mathrm{a}}\\
         x_1^{\mathrm{v}}
    \end{bmatrix}
    +
    (1-\tau)
    \begin{bmatrix}
        x_0^{\mathrm{a}}\\
        x_0^{\mathrm{v}}
    \end{bmatrix}.
\end{equation}
The corresponding ground-truth velocity is given by
\begin{equation}
    \nu_{\tau}
    =
    \frac{d x_{\tau}}{d\tau}
    =
    \begin{bmatrix}
        x_1^{\mathrm{a}} - x_0^{\mathrm{a}} \\
        x_1^{\mathrm{v}} - x_0^{\mathrm{v}}
    \end{bmatrix}.
\end{equation}
Finally, \textit{GigaWorld-Policy-0.5} is optimized by regressing the model-predicted velocity field to the ground-truth flow velocity:
\begin{equation}
    \mathcal{L}
    =
    \mathbb{E}_{x_1,x_0,\tau}
    \left[
    \left\|
    g_{\theta}
    \left(
        x_{\tau}, \tau \mid o_t^{\mathrm{comp}}, s_t, l
    \right)
    -
    \nu_{\tau}
    \right\|_2^2
    \right].
\end{equation}

\subsection{Model Architecture}

Fig.~\ref{fig:framework} illustrates the MoT-based architecture of \textit{GigaWorld-Policy-0.5}. The model takes visual observations, proprioceptive states, action chunks, and language instructions as inputs. For visual inputs, we use the visual VAE from Wan~\citep{wan} to encode the composite observations $o^{\mathrm{comp}}$ into visual latent tokens. For non-visual inputs, robot states and action chunks are mapped to the visual hidden dimension via multi-layer perceptrons (MLPs), producing state tokens and action tokens. The language instruction is encoded by umT5~\citep{umt5}, whose text embeddings are used as conditioning signals for instruction-following control.

Given these tokens, \textit{GigaWorld-Policy-0.5} follows the action-centered world action framework but replaces the fully shared Transformer backbone with an MoT structure. The MoT architecture separates the two central modeling components of action-centered WAMs, namely visual dynamics modeling and action generation, into a visual expert and an action expert. The visual expert processes current and future visual tokens and models scene evolution in the latent visual space, while the action expert focuses on action-token denoising and action-chunk prediction. Each expert is equipped with its own cross-attention and feed-forward network (FFN) modules, enabling modality-specific language conditioning and nonlinear transformation. These two experts are connected through multi-modal self-attention, enabling information exchange across visual, state, and action tokens while preserving a unified token sequence for joint world action modeling.

In particular, the multi-modal self-attention module follows GigaWorld-Policy~\citep{gigaworldpolicy} by adopting an action-centered causal mask to regulate cross-modal information flow. Specifically, action tokens are allowed to attend to the current visual tokens, state tokens, and language conditioning, but are prevented from attending to future visual tokens. Future visual tokens, on the other hand, can attend to the current context and action tokens, enabling action-conditioned future visual prediction. This causal attention mask prevents information leakage from future observations into action prediction, while allowing future visual dynamics to provide dense training-time supervision. At inference time, future visual tokens can be omitted, and the model directly decodes action tokens for low-latency control.

For parameter initialization, \textit{GigaWorld-Policy-0.5} initializes the visual expert with pretrained visual weights from GigaWorld-1~\citep{gigaworld1}, a large-scale pretrained world model. This provides the visual expert with a world-model prior for visual dynamics modeling, allowing the model to inherit general knowledge about scene evolution before being adapted to robot-specific observations and interactions. 
For the newly introduced action expert, we initialize its parameters from the corresponding visual-expert weights. For parameters whose dimensions are not aligned, we initialize the action-expert parameters by taking the leading $n$ dimensions of the corresponding visual-expert weights, where $n$ denotes the target dimensionality of the action-expert parameter.

\subsection{Training Pipeline}

The training pipeline of \textit{GigaWorld-Policy-0.5} consists of two stages: robot-data pretraining and target-robot post-training. Before these two stages, the visual expert is initialized from GigaWorld-1~\citep{gigaworld1}, which is pretrained on over ten thousand hours of video data. This initialization provides \textit{GigaWorld-Policy-0.5} with a general world-model prior for visual dynamics modeling.

In the pretraining stage, we train \textit{GigaWorld-Policy-0.5} on 2K hours of filtered open-source robot data~\citep{bu2025agibot,liu2025rdt,wu2024robomind,tan2025anypos} and internally collected real-robot data. This stage adapts the pretrained world-model prior to robot-centric scenarios, including robot embodiments, manipulation workspaces, camera viewpoints, and interaction patterns. The model learns to predict future visual evolution under robot-relevant observations and actions, improving its alignment with robotic control settings before target-domain specialization. Notably, to better model the relationship between actions and visual dynamics, we also incorporate action-conditioned world-model training during this stage, where future visual evolution is predicted under robot-relevant observations and actions. This encourages the model to learn how robot actions affect scene changes before target-domain specialization.

In the post-training stage, \textit{GigaWorld-Policy-0.5} is trained on target real-robot trajectories that contain aligned observations, language instructions, robot states, and action sequences. The model jointly optimizes action prediction and future visual dynamics modeling under the action-centered causal structure. The action objective teaches the model to generate instruction-conditioned action chunks for closed-loop control, while the future-visual objective provides auxiliary supervision that encourages action representations to remain consistent with plausible scene evolution. During deployment, we use the action-only inference path: future visual tokens are not generated, and the model directly decodes actions conditioned on the current observation, state, and instruction.

\begin{figure}[!t]
\centering
\captionsetup{type=figure, justification=justified, singlelinecheck=true}
\includegraphics[width=\textwidth]{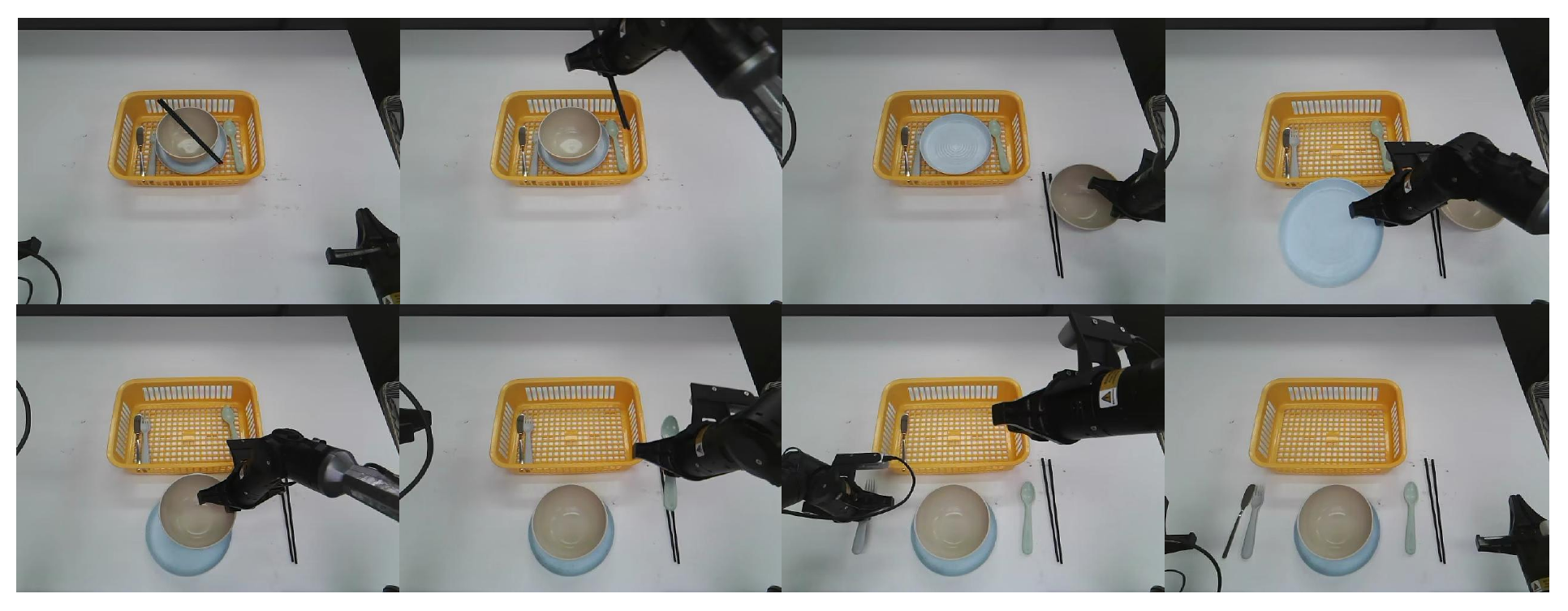}
\caption{
Real-world demonstration of \textit{GigaWorld-Policy-0.5} on the \textit{Tableware Arrangement} task.
}
\label{fig:demo-canjv}
\end{figure}

\subsection{Inference Acceleration}

To support real-world deployment on edge computing platforms, we optimize the inference stack through KV caching, graph compilation, and a lightweight C\texttt{++} runtime. During autoregressive action generation, the visual and language context remains unchanged across decoding steps; we therefore cache the key-value states of all attention layers after encoding the input context and reuse them for subsequent action-token predictions, avoiding redundant attention computation. We further apply \texttt{torch.compile} to optimize the inference graph, which reduces Python-level dispatch overhead and enables backend-level optimizations such as operator fusion and more efficient memory scheduling. Finally, we deploy the optimized policy in a C\texttt{++} runtime that integrates image preprocessing, tensor construction, model execution, KV-cache management, and action post-processing into a unified native pipeline. This design removes the Python dependency, reduces runtime overhead, and simplifies integration with robot-control middleware, enabling low-latency closed-loop policy execution on edge devices.
\begin{figure}[!th]
\centering
\captionsetup{type=figure, justification=justified, singlelinecheck=true}
\includegraphics[width=\textwidth]{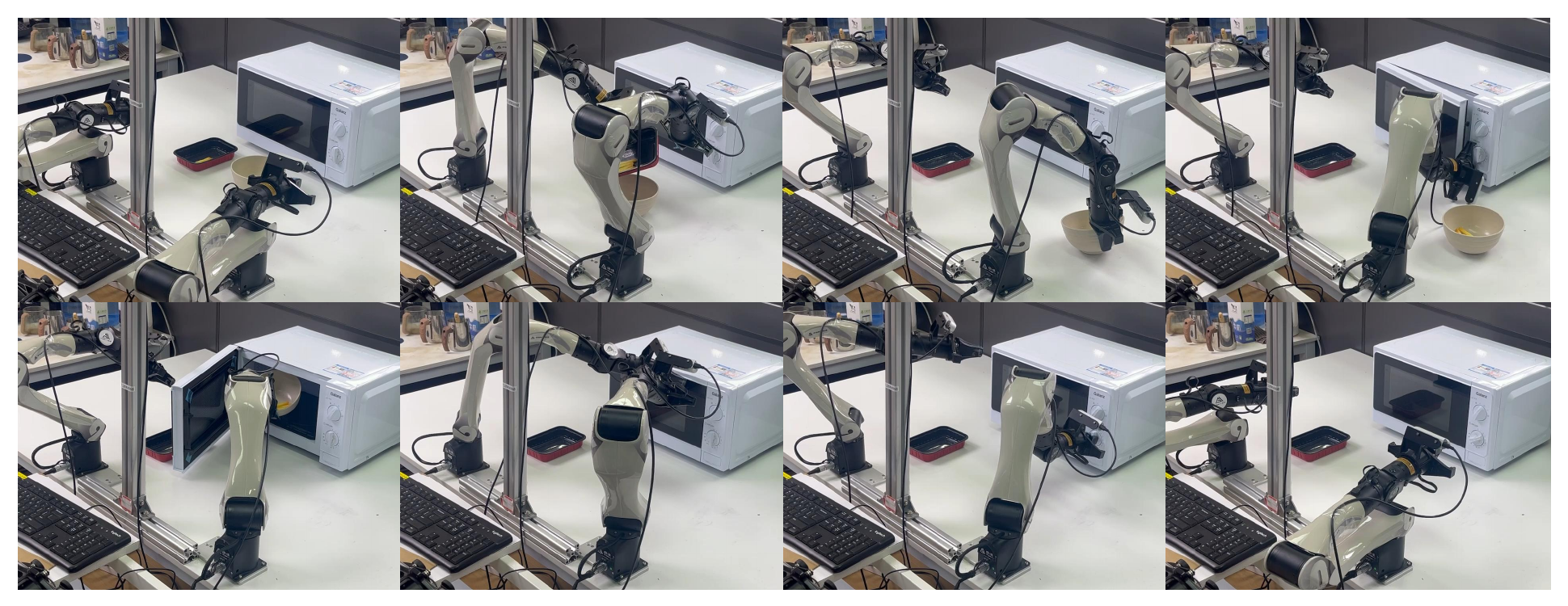}
\caption{
Real-world demonstration of \textit{GigaWorld-Policy-0.5} on the \textit{Food Heating} task.
}
\label{fig:demo-heatfood}
\end{figure}

\section{Experiments}

\noindent{\textbf{Evaluation Metrics.}} We primarily report the Success Rate (SR) to evaluate task completion performance. For real-world experiments, we evaluate gripper-based manipulation under two settings: \textbf{text following} and \textbf{long-horizon task} execution. For \textbf{text following}, we use a graded SR to better capture partial task completion: each trial is scored by four equally weighted stages, with 0.25 assigned to each stage, including reaching the target object, grasping it, moving to the target location, and successfully placing it. For \textbf{long-horizon tasks}, SR follows the same binary definition as in simulation, where SR = 1 only if the full task sequence is completed and SR = 0 otherwise.

\noindent{\textbf{Baselines.}} We compare \textit{GigaWorld-Policy-0.5} with several representative baselines spanning two major paradigms. For VLM-based VLA methods, we include $\pi_{0.5}$~\citep{pi0.5}, which uses a large vision-language backbone to align visual observations with language instructions and decode low-level control actions through an action head. For WAM-based methods, we include Motus~\citep{motus}, FastWAM~\citep{fastwam}, and GigaWorld-Policy~\citep{gigaworldpolicy}, which incorporate visual dynamics or world-modeling objectives to support policy learning.

\noindent{\textbf{Implementation Details.}} We use the pretrained GigaWorld-1~\citep{gigaworld1} to initialize the visual expert, enabling the model to inherit strong visual generative representations. The action expert is then initialized from the visual expert by copying the corresponding weights; for layers with mismatched dimensions, we retain the compatible submatrices and truncate the remaining entries to match the action-expert configuration. In our implementation, the visual expert uses a hidden dimension of 3072 and an FFN dimension of 14336, while the action expert is made more lightweight with a hidden dimension of 1024 and an FFN dimension of 4096. For the action chunk length and future-observation stride, we follow the settings adopted in GigaWorld-Policy.

\begin{table}[h]
\centering
\captionsetup{type=table, justification=justified, singlelinecheck=true}
\caption{Evaluation of text-following ability on the fruit-picking task.}
\label{tab:text_following_fruit}
\resizebox{\textwidth}{!}{
\begin{tabular}{llccccc}
\toprule
Task Name & Text Instruction & $\pi_{0.5}$~\citep{pi0.5} & Motus~\citep{motus} & FastWAM~\citep{fastwam} & GigaWorld-Policy~\citep{gigaworldpolicy} & \textit{GigaWorld-Policy-0.5} \\
\midrule
\multirow{6}{*}{Pick the Fruit}
& Pick the \textit{banana} and place it into the basket.     & 0.88 & 0.93 & 0.83 & 0.90 & \textbf{0.95} \\
& Pick the \textit{apple} and place it into the basket.      & 0.73 & 0.78 & 0.75 & 0.73 & \textbf{0.80} \\
& Pick the \textit{lemon} and place it into the basket.      & 0.68 & 0.75 & 0.73 & 0.78 & \textbf{0.83} \\
& Pick the \textit{grape} and place it into the basket.      & 0.85 & 0.83 & 0.88 & 0.88 & \textbf{0.93} \\
& Pick the \textit{avocado} and place it into the basket.    & 0.68 & 0.70 & 0.75 & 0.73 & \textbf{0.78} \\
& Pick the \textit{strawberry} and place it into the basket. & 0.78 & 0.80 & 0.73 & 0.78 & \textbf{0.85} \\
\midrule
Average & -- & 0.76 & 0.80 & 0.78 & 0.80 & \textbf{0.85} \\
\bottomrule
\end{tabular}
}
\end{table}

\subsection{Real-World Results}

We conduct gripper-based manipulation experiments on an AgileX PiPER 6-DoF robotic arm from two complementary perspectives. \textbf{Text following} evaluates language grounding by executing multiple instructions under the same scene configuration. Each trial is scored using four equally weighted stages (0.25 point each): reaching the target object, grasping the target object, moving to the target placement location, and successfully placing the object. The final score is averaged over 10 real-world trials. \textbf{Long-horizon tasks} evaluate end-to-end sequential manipulation, where each task is executed 10 times on the real robot and only the overall task success rate is reported.

Tab.~\ref{tab:text_following_fruit} and~\ref{tab:text_following_object_placement} evaluate text-following ability under variations in object semantics and target locations. On the fruit-picking task, \textit{GigaWorld-Policy-0.5} achieves an average success rate of 0.85, outperforming $\pi_{0.5}$~\citep{pi0.5}, Motus~\citep{motus}, FastWAM~\citep{fastwam}, and GigaWorld-Policy~\citep{gigaworldpolicy} by 0.09, 0.05, 0.07, and 0.05, respectively. It consistently obtains the highest success rate across all six fruit categories, with particularly clear gains on lemon and avocado, suggesting that the model can reliably ground fine-grained language descriptions to visually distinct objects. On the more compositional object-placement task, \textit{GigaWorld-Policy-0.5} achieves an average success rate of 0.89, outperforming the strongest baseline, Motus, by 0.06. This task requires jointly identifying the referenced object and following the specified destination relation (e.g., placing an object \emph{on} a plate or \emph{into} a basket). Our method attains the best result on every instruction and shows notable improvements for bowl-to-basket and fork-to-basket instructions, indicating stronger grounding of both object identities and spatial goal specifications. Overall, the results demonstrate that \textit{GigaWorld-Policy-0.5} follows diverse natural-language instructions more reliably than prior methods, particularly when task execution requires precise object selection and compositional object-destination reasoning.

\begin{table}[h]
\centering
\captionsetup{type=table, justification=justified, singlelinecheck=true}
\caption{Evaluation of text-following ability on the object-placement task.}
\label{tab:text_following_object_placement}
\resizebox{\textwidth}{!}{
\begin{tabular}{llccccc}
\toprule
Task Name & Text Instruction & $\pi_{0.5}$~\citep{pi0.5} & Motus~\citep{motus} & FastWAM~\citep{fastwam} & GigaWorld-Policy~\citep{gigaworldpolicy}  & \textit{GigaWorld-Policy-0.5} \\
\midrule
\multirow{6}{*}{Place Objects}
& Pick up the \textit{bowl} and place it on the \textit{plate}.      & 0.87 & 0.88 & 0.73 & 0.80 & \textbf{0.93} \\
& Pick up the \textit{fork} and place it on the \textit{plate}.      & 0.78 & 0.83 & 0.85 & 0.80 & \textbf{0.88} \\
& Pick up the \textit{spoon} and place it on the \textit{plate}.     & 0.73 & 0.83 & 0.80 & 0.75 & \textbf{0.85} \\
& Pick up the \textit{bowl} and place it into the \textit{basket}.   & 0.75 & 0.83 & 0.80 & 0.88 & \textbf{0.95} \\
& Pick up the \textit{fork} and place it into the \textit{basket}.   & 0.65 & 0.75 & 0.68 & 0.78 & \textbf{0.85} \\
& Pick up the \textit{spoon} and place it into the \textit{basket}.  & 0.80 & 0.85 & 0.75 & 0.83 & \textbf{0.88} \\
\midrule
Average & -- & 0.76 & 0.83 & 0.77 & 0.81 & \textbf{0.89} \\
\bottomrule
\end{tabular}
}
\end{table}

Table~\ref{tab:long_horizon_tasks} reports end-to-end success rates on three long-horizon manipulation tasks, each of which requires the policy to execute multiple temporally dependent steps. \textit{GigaWorld-Policy-0.5} achieves the best performance on every task, attaining an average success rate of 0.80. Compared with the strongest baseline, it yields an absolute improvement of 0.20 and a relative improvement of 33\%. This substantial gain highlights the effectiveness of our method in long-horizon task execution, where successful completion depends on maintaining coherent action sequences and accurately handling intermediate state transitions. The results further suggest that the predictive representations acquired through world modeling provide useful temporal and interaction-aware information for long-horizon robotic manipulation.

\begin{table}[h]
\centering
\captionsetup{type=table, justification=justified, singlelinecheck=true}
\caption{Evaluation of end-to-end success rates on long-horizon tasks.}
\label{tab:long_horizon_tasks}
\resizebox{\textwidth}{!}{
\begin{tabular}{lccccc}
\toprule
Task Name & $\pi_{0.5}$~\citep{pi0.5} & Motus~\citep{motus} & FastWAM~\citep{fastwam} & GigaWorld-Policy~\citep{gigaworldpolicy} & \textit{GigaWorld-Policy-0.5} \\
\midrule
Food Heating  & 0.50 & 0.60 & 0.50 & 0.60 & \textbf{0.80} \\
Tableware Arrangement   & 0.70 & 0.60 & 0.50 & 0.60 & \textbf{0.80} \\
\midrule
Average & 0.60 & 0.60 & 0.50 & 0.60 & \textbf{0.80}\\
\bottomrule
\end{tabular}}
\end{table}

\subsection{Ablation Studies}

We conduct two groups of ablation studies to analyze \textit{GigaWorld-Policy-0.5}. The first group focuses on standard component ablations, where we isolate the contribution of key design choices in the model and training pipeline. The second group uses the AutoResearch~\cite{karpathy2026autoresearch} pipeline to study the effect of hyperparameter choices, such as learning rate and warmup strategy, under a unified evaluation protocol.

\paragraph{Effect of the mixed AC-WM and WAM pretraining.}
We evaluate whether incorporating Action-Conditioned World-Model (AC-WM) training during pretraining improves downstream policy learning. This ablation compares two pretraining settings: one using standard WAM pretraining only, and the other mixing AC-WM and WAM pretraining, where future visual observations are predicted conditioned on robot actions. After pretraining, both models are post-trained using the same 

\begin{minipage}[t]{0.5\textwidth}
policy training recipe and evaluated on the \textit{pick the fruit} task. 

Fig.~\ref{fig:ablation_acwm} shows that the model pretrained with mixed AC-WM and WAM outperforms the baseline throughout post-training: it converges faster, attains higher success rates, and ultimately reaches a success rate of 0.85. Notably, strong performance emerges at substantially earlier training steps, indicating that explicitly modeling how robot actions drive visual state transitions yields more transferable action representations. As a result, downstream policy learning becomes more sample-efficient, requiring fewer post-training updates to achieve competitive closed-loop control performance.
\end{minipage}\hfill
\begin{minipage}[t]{0.48\textwidth}
  \centering
  \vspace{-0.4em}
  \captionsetup{type=figure, justification=justified, singlelinecheck=false}
  \includegraphics[width=\linewidth]{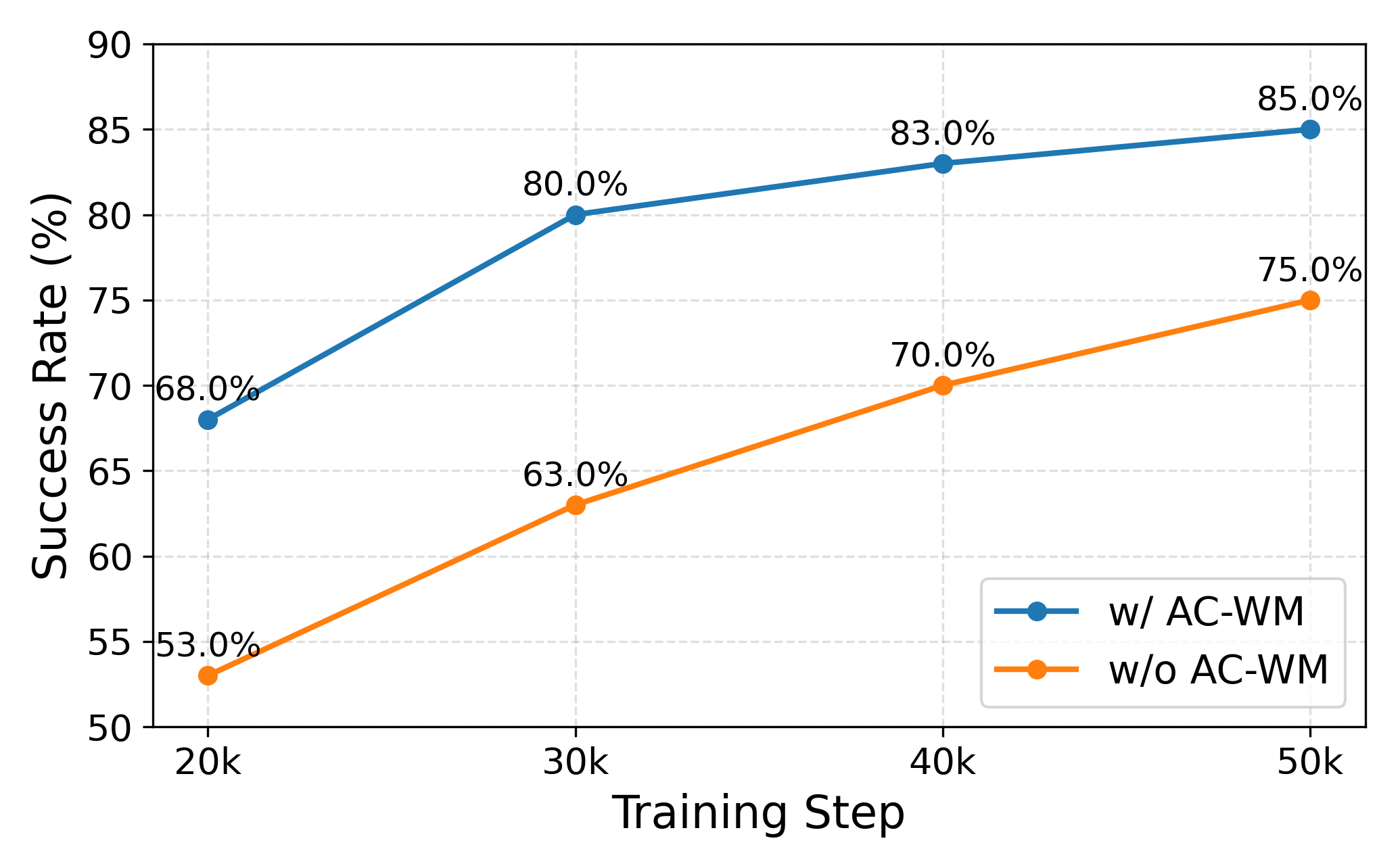}
  \caption{Success rates at different training steps in the AC-WM ablation study.}
  \label{fig:ablation_acwm}
\end{minipage}
\vspace{-0.5em}

\paragraph{Effect of MoT architecture.}

We evaluate the effect of the MoT architecture on inference efficiency. The MoT architecture processes action generation and video dynamics modeling with separate experts. In our design, we reduce the dimensional configuration of the action expert relative to the video expert, making the action pathway more lightweight than the video modeling pathway. As a result, although \textit{GigaWorld-Policy-0.5} introduces additional parameters due to the expert-based architecture, the computation required for action-only inference is reduced. In addition, this expert-separated design makes it easier to initialize the video expert from pretrained video generation models, which provides stronger visual dynamics representations and further improves downstream policy performance.

As shown in Tab.~\ref{tab:mot_efficiency}, this architectural design leads to clear inference efficiency gains. With KV cache and torch compilation enabled, \textit{GigaWorld-Policy-0.5} achieves an inference latency of 189 ms on an A100 GPU. Compared with FastWAM under the same torch-compiled setting, \textit{GigaWorld-Policy-0.5} reduces latency from 229 ms to 189 ms, corresponding to a 17.5\% speedup. It also outperforms the VLA-based $\pi_{0.5}$~\citep{pi0.5} in inference efficiency, reducing latency from 225 ms to 189 ms. Beyond the A100 setting, \textit{GigaWorld-Policy-0.5} further achieves 110 ms latency on a local RTX 4090 edge-side setup, matching the inference speed of $\pi_{0.5}$. With a C\texttt{++} deployment, latency can be further reduced to 85 ms, 23\% faster than $\pi_{0.5}$ and 53\% faster than FastWAM~\citep{fastwam}, demonstrating its potential for efficient real-world deployment.

\begin{table}[t]
\centering
\captionsetup{type=table, justification=justified, singlelinecheck=true}
\caption{Comparison of inference efficiency and real-robot performance.}
\label{tab:mot_efficiency}
\begin{tabular}{lccc}
\toprule
Method & Latency on A100 (ms) $\downarrow$ & Latency on RTX 4090 (ms) $\downarrow$ & SR on Real-Robot $\uparrow$\\
\midrule
$\pi_{0.5}$~\citep{pi0.5}     & 225 & 110  & 0.76 \\
Motus~\citep{motus}           & 3231 & - & 0.80 \\
FastWAM~\citep{fastwam}       & 229 & 182  & 0.78 \\
GigaWorld-Policy~\citep{gigaworldpolicy} & 360 & 293  & 0.80 \\
\textit{GigaWorld-Policy-0.5} & 189 & 110 & \textbf{0.85} \\
\ w/ C\texttt{++} deployment & \textbf{140} & \textbf{85} & \textbf{0.85} \\
\bottomrule
\end{tabular}
\end{table}

\begin{figure}[ht]
\centering
\captionsetup{type=figure, justification=justified, singlelinecheck=false}
\includegraphics[width=\textwidth]{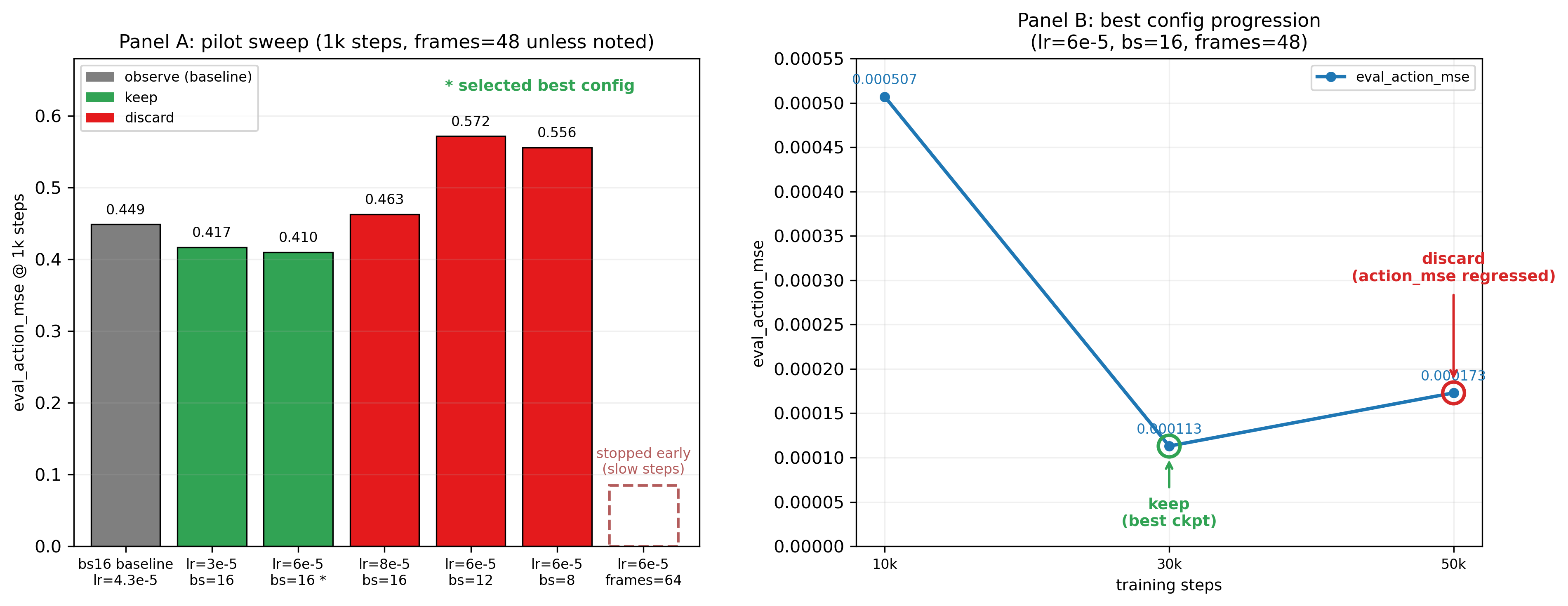}
\caption{
AutoResearch hyperparameter search and training progression on the \textit{pick the fruit} task. Left: 1K-step pilot sweep over learning rate and batch-size configurations, where green bars indicate retained candidates and red bars indicate discarded configurations. Right: extended training progression under the selected configuration, showing that the model achieves the best validation action MSE at 30K steps.
}
\label{fig:autoresearch}
\end{figure}

\paragraph{AutoResearch-driven hyperparameter study.}
We use AutoResearch~\citep{karpathy2026autoresearch} to systematically explore the training hyperparameters of \textit{GigaWorld-Policy-0.5}, including the learning rate and warmup schedule. AutoResearch automates the full optimization loop, including candidate configuration generation, pilot training, validation metric collection, candidate selection, and extended training. This procedure is particularly useful for \textit{GigaWorld-Policy-0.5}, as the model jointly optimizes action prediction and future visual dynamics modeling, and different hyperparameter choices may affect these two objectives in different ways.

We conduct the study on the \textit{pick the fruit} task using approximately 3.9 hours of robot demonstration data. The dataset contains 930 episodes, of which 300 episodes are used for training and 30 episodes are held out for validation. All candidate runs use the same model initialization, data split, optimizer, and evaluation protocol. To improve the efficiency of hyperparameter exploration, AutoResearch first uses short 1K-step pilot runs to rapidly iterate over different candidate configurations. After identifying promising hyperparameters, AutoResearch performs longer training runs under the selected configuration for final model selection.

\begin{table}[h]
\centering
\caption{AutoResearch learning-rate sweep. Each candidate is trained for 1K steps under the same setting. }
\label{tab:autoresearch_lr}
\begin{tabular}{lccc}
\toprule
Learning Rate & Train Action Loss $\downarrow$ & Train Visual Loss $\downarrow$ & Eval Action MSE $\downarrow$ \\
\midrule
$3\times10^{-5}$ & 0.257300 & \textbf{0.172330} & 0.416832\\
$4.316\times10^{-5}$ & 0.256593 & 0.172893 & 0.449387\\
$6\times10^{-5}$  & \textbf{0.252476} & 0.173318 & \textbf{0.409764}\\
$8\times10^{-5}$  & 0.261832 & 0.175986 & 0.461381\\
\bottomrule
\end{tabular}
\end{table}

As shown on the left side of Fig.~\ref{fig:autoresearch}, AutoResearch first sweeps over candidate learning rates using these 1K-step pilot runs. The detailed learning-rate ablation results are reported in Tab.~\ref{tab:autoresearch_lr}: among the tested candidates, $6\times10^{-5}$ achieves the lowest training action loss of 0.252476 and the best evaluation action MSE of 0.409764, while $3\times10^{-5}$ obtains the lowest training visual loss of 0.172330. Since action prediction quality is more directly related to downstream policy execution, AutoResearch selects $6\times10^{-5}$ as the learning rate for the next-stage search.

AutoResearch then fixes this learning rate and conducts an additional batch-size sweep. The batch-size search shows that alternative batch sizes do not outperform the original setting, so AutoResearch retains the original batch size of 16. Based on these results, AutoResearch selects $6\times10^{-5}$ and a batch size of 16 as the final hyperparameter configuration. After fixing these hyperparameters, AutoResearch further scales up the training by extending the number of training steps. As shown on the right side of Fig.~\ref{fig:autoresearch}, the model reaches its best validation performance at 30K training steps. We therefore use the 30K-step checkpoint for subsequent real-robot evaluation.

\section{Conclusion}

In this technical report, we introduced GigaWorld-Policy-0.5, an enhanced action-centered World Action Model built upon GigaWorld-Policy. GigaWorld-Policy-0.5 preserves the key principle of using future visual dynamics as training-time supervision while enabling action-only decoding during inference. On top of this formulation, the model adopts a Mixture-of-Transformers architecture to separate visual dynamics modeling from action generation, reducing active computation for low-latency deployment. We also introduce a mixed pretraining strategy that combines action-conditioned world modeling with WAM training, encouraging the model to better capture action-induced visual changes and improving action modeling. Finally, an agent-based AutoResearch pipeline is used to make hyperparameter search more systematic and efficient. Overall, GigaWorld-Policy-0.5 demonstrates a practical path toward faster and more deployable World Action Models for robot control. With C\texttt{++} deployment on a local RTX~4090, it further achieves 85 ms inference latency, improving the efficiency-performance trade-off of action-centered WAMs by retaining dense supervision from future visual dynamics while reducing the cost of action generation at inference time.

\clearpage
\setcitestyle{numbers}
\bibliographystyle{plainnat}
\bibliography{main}

@String(AAAI  = {AAAI})

@article{gigaworld0,
  title={Gigaworld-0: World models as data engine to empower embodied ai},
  author={Team, GigaWorld and Ye, Angen and Wang, Boyuan and Ni, Chaojun and Huang, Guan and Zhao, Guosheng and Li, Haoyun and Zhu, Jiagang and Li, Kerui and Xu, Mengyuan and others},
  journal={arXiv preprint arXiv:2511.19861},
  year={2025}
}

@article{wan,
  title={Wan: Open and advanced large-scale video generative models},
  author={Wan, Team and Wang, Ang and Ai, Baole and Wen, Bin and Mao, Chaojie and Xie, Chen-Wei and Chen, Di and Yu, Feiwu and Zhao, Haiming and Yang, Jianxiao and others},
  journal={arXiv preprint arXiv:2503.20314},
  year={2025}
}

@inproceedings{drivedreamer,
  title={Drivedreamer: Towards real-world-drive world models for autonomous driving},
  author={Wang, Xiaofeng and Zhu, Zheng and Huang, Guan and Chen, Xinze and Zhu, Jiagang and Lu, Jiwen},
  booktitle={European conference on computer vision},
  pages={55--72},
  year={2024},
  organization={Springer}
}

@inproceedings{drivedreamer2,
  title={Drivedreamer-2: Llm-enhanced world models for diverse driving video generation},
  author={Zhao, Guosheng and Wang, Xiaofeng and Zhu, Zheng and Chen, Xinze and Huang, Guan and Bao, Xiaoyi and Wang, Xingang},
  booktitle={Proceedings of the AAAI Conference on Artificial Intelligence},
  volume={39},
  number={10},
  pages={10412--10420},
  year={2025}
}

@article{cosmos,
  title={Cosmos world foundation model platform for physical ai},
  author={Agarwal, Niket and Ali, Arslan and Bala, Maciej and Balaji, Yogesh and Barker, Erik and Cai, Tiffany and Chattopadhyay, Prithvijit and Chen, Yongxin and Cui, Yin and Ding, Yifan and others},
  journal={arXiv preprint arXiv:2501.03575},
  year={2025}
}

@article{cosmos3,
  title={Cosmos 3: Omnimodal world models for physical ai},
  author={Agarwal, Niket and Ali, Arslan and Allen, Jon and Antolini, Martin and Aubame, Adeline and Azzolini, Alisson and Bai, Junjie and Bala, Maciej and Balaji, Yogesh and Bapst, Josh and others},
  journal={arXiv preprint arXiv:2606.02800},
  year={2026}
}

@article{worldmodel,
  title={A path towards autonomous machine intelligence version 0.9. 2, 2022-06-27},
  author={LeCun, Yann},
  journal={Open Review},
  year={2022}
}

@article{pi0.7,
  title={$\pi_{0.7}$: a Steerable Generalist Robotic Foundation Model with Emergent Capabilities},
  author={Intelligence, Physical and Ai, Bo and Amin, Ali and Aniceto, Raichelle and Balakrishna, Ashwin and Balke, Greg and Black, Kevin and Bokinsky, George and Cao, Shihao and Charbonnier, Thomas and others},
  journal={arXiv preprint arXiv:2604.15483},
  year={2026}
}

@article{wog,
  title={World guidance: World modeling in condition space for action generation},
  author={Su, Yue and Chen, Sijin and Shi, Haixin and Liu, Mingyu and Zhang, Zhengshen and Huang, Ningyuan and Zhong, Weiheng and Zhu, Zhengbang and Liu, Yuxiao and Liu, Xihui},
  journal={arXiv preprint arXiv:2602.22010},
  year={2026}
}

@article{beingh0.7,
  title={Being-h0. 7: A latent world-action model from egocentric videos},
  author={Luo, Hao and Zhang, Wanpeng and Feng, Yicheng and Zheng, Sipeng and Xu, Haiweng and Xu, Chaoyi and Xi, Ziheng and Fu, Yuhui and Lu, Zongqing},
  journal={arXiv preprint arXiv:2605.00078},
  year={2026}
}

@article{worldvla,
  title={Worldvla: Towards autoregressive action world model},
  author={Cen, Jun and Yu, Chaohui and Yuan, Hangjie and Jiang, Yuming and Huang, Siteng and Guo, Jiayan and Li, Xin and Song, Yibing and Luo, Hao and Wang, Fan and others},
  journal={arXiv preprint arXiv:2506.21539},
  year={2025}
}

@article{gigabrain0,
  title={Gigabrain-0: A world model-powered vision-language-action model},
  author={Team, GigaBrain and Ye, Angen and Wang, Boyuan and Ni, Chaojun and Huang, Guan and Zhao, Guosheng and Li, Haoyun and Li, Jie and Zhu, Jiagang and Feng, Lv and others},
  journal={arXiv preprint arXiv:2510.19430},
  year={2025}
}

@article{gigabrain0.5m,
  title={Gigabrain-0.5 m*: a vla that learns from world model-based reinforcement learning},
  author={Team, GigaBrain and Wang, Boyuan and Li, Bohan and Ni, Chaojun and Huang, Guan and Zhao, Guosheng and Li, Hao and Li, Jie and Lv, Jindi and Liu, Jingyu and others},
  journal={arXiv preprint arXiv:2602.12099},
  year={2026}
}

@inproceedings{swiftvla,
  title={Swiftvla: Unlocking spatiotemporal dynamics for lightweight vla models at minimal overhead},
  author={Ni, Chaojun and Chen, Cheng and Wang, Xiaofeng and Zhu, Zheng and Zheng, Wenzhao and Wang, Boyuan and Chen, Tianrun and Zhao, Guosheng and Li, Haoyun and Dong, Zhehao and others},
  booktitle={Proceedings of the IEEE/CVF Conference on Computer Vision and Pattern Recognition},
  pages={13474--13485},
  year={2026}
}

@article{pi0,
  title={$\pi_{0}$: A Vision-Language-Action Flow Model for General Robot Control},
  author={Black, Kevin and Brown, Noah and Driess, Danny and Esmail, Adnan and Equi, Michael and Finn, Chelsea and Fusai, Niccolo and Groom, Lachy and Hausman, Karol and Ichter, Brian and others},
  journal={arXiv preprint arXiv:2410.24164},
  year={2024}
}

@article{pi0.5,
  title={$\pi_{0.5}$: a Vision-Language-Action Model with Open-World Generalization},
  author={Intelligence, Physical and Black, Kevin and Brown, Noah and Darpinian, James and Dhabalia, Karan and Driess, Danny and Esmail, Adnan and Equi, Michael and Finn, Chelsea and Fusai, Niccolo and others},
  journal={arXiv preprint arXiv:2504.16054},
  year={2025}
}

@article{pi0.6,
  title={$\pi_{0.6}$: a VLA That Learns From Experience},
  author={Intelligence, Physical and Amin, Ali and Aniceto, Raichelle and Balakrishna, Ashwin and Black, Kevin and Conley, Ken and Connors, Grace and Darpinian, James and Dhabalia, Karan and DiCarlo, Jared and others},
  journal={arXiv preprint arXiv:2511.14759},
  year={2025}
}

@article{gr00t,
  title={Gr00t n1: An open foundation model for generalist humanoid robots},
  author={Bjorck, Johan and Casta{\~n}eda, Fernando and Cherniadev, Nikita and Da, Xingye and Ding, Runyu and Fan, Linxi and Fang, Yu and Fox, Dieter and Hu, Fengyuan and Huang, Spencer and others},
  journal={arXiv preprint arXiv:2503.14734},
  year={2025}
}

@article{walloss,
  title={Igniting vlms toward the embodied space},
  author={Zhai, Andy and Liu, Brae and Fang, Bruno and Cai, Chalse and Ma, Ellie and Yin, Ethan and Wang, Hao and Zhou, Hugo and Wang, James and Shi, Lights and others},
  journal={arXiv preprint arXiv:2509.11766},
  year={2025}
}

@inproceedings{motus,
  title={Motus: A unified latent action world model},
  author={Bi, Hongzhe and Tan, Hengkai and Xie, Shenghao and Wang, Zeyuan and Huang, Shuhe and Liu, Haitian and Zhao, Ruowen and Feng, Yao and Xiang, Chendong and Rong, Yinze and others},
  booktitle={Proceedings of the IEEE/CVF Conference on Computer Vision and Pattern Recognition},
  pages={35101--35113},
  year={2026}
}

@article{motubrain,
  title={Motubrain: An advanced world action model for robot control},
  author={Team, MotuBrain and Xiang, Chendong and Bao, Fan and Liu, Haitian and Tan, Hengkai and Bi, Hongzhe and Li, James and Liu, Jiabao and Pang, Jingrui and Jing, Kiro and others},
  journal={arXiv preprint arXiv:2604.27792},
  year={2026}
}

@article{lingbotva,
  title={Causal World Modeling for Robot Control},
  author={Li, Lin and Zhang, Qihang and Luo, Yiming and Yang, Shuai and Wang, Ruilin and Han, Fei and Yu, Mingrui and Gao, Zelin and Xue, Nan and Zhu, Xing and others},
  journal={arXiv preprint arXiv:2601.21998},
  year={2026}
}

@article{lingbotvla,
  title={A Pragmatic VLA Foundation Model},
  author={Wu, Wei and Lu, Fan and Wang, Yunnan and Yang, Shuai and Liu, Shi and Wang, Fangjing and Zhu, Qian and Sun, He and Wang, Yong and Ma, Shuailei and others},
  journal={arXiv preprint arXiv:2601.18692},
  year={2026}
}

@article{videovla,
  title={Videovla: Video generators can be generalizable robot manipulators},
  author={Shen, Yichao and Wei, Fangyun and Du, Zhiying and Liang, Yaobo and Lu, Yan and Yang, Jiaolong and Zheng, Nanning and Guo, Baining},
  journal={Advances in neural information processing systems},
  volume={38},
  pages={95597--95621},
  year={2026}
}

@article{dreamzero,
  title={World action models are zero-shot policies},
  author={Ye, Seonghyeon and Ge, Yunhao and Zheng, Kaiyuan and Gao, Shenyuan and Yu, Sihyun and Kurian, George and Indupuru, Suneel and Tan, You Liang and Zhu, Chuning and Xiang, Jiannan and others},
  journal={arXiv preprint arXiv:2602.15922},
  year={2026}
}

@article{fastwam,
  title={Fast-wam: Do world action models need test-time future imagination?},
  author={Yuan, Tianyuan and Dong, Zibin and Liu, Yicheng and Zhao, Hang},
  journal={arXiv preprint arXiv:2603.16666},
  year={2026}
}

@article{gigaworldpolicy,
  title={GigaWorld-Policy: An Efficient Action-Centered World--Action Model},
  author={Ye, Angen and Wang, Boyuan and Ni, Chaojun and Huang, Guan and Zhao, Guosheng and Li, Hao and Li, Hengtao and Li, Jie and Lv, Jindi and Liu, Jingyu and others},
  journal={arXiv preprint arXiv:2603.17240},
  year={2026}
}

@article{du2023learning,
  title={Learning universal policies via text-guided video generation},
  author={Du, Yilun and Yang, Sherry and Dai, Bo and Dai, Hanjun and Nachum, Ofir and Tenenbaum, Josh and Schuurmans, Dale and Abbeel, Pieter},
  journal={Advances in neural information processing systems},
  volume={36},
  pages={9156--9172},
  year={2023}
}

@inproceedings{wu2024unleashing,
  title={Unleashing large-scale video generative pre-training for visual robot manipulation},
  author={Wu, Hongtao and Jing, Ya and Cheang, Chilam and Chen, Guangzeng and Xu, Jiafeng and Li, Xinghang and Liu, Minghuan and Li, Hang and Kong, Tao},
  booktitle={International Conference on Learning Representations},
  volume={2024},
  pages={10641--10662},
  year={2024}
}

@article{zhu2025unified,
  title={Unified world models: Coupling video and action diffusion for pretraining on large robotic datasets},
  author={Zhu, Chuning and Yu, Raymond and Feng, Siyuan and Burchfiel, Benjamin and Shah, Paarth and Gupta, Abhishek},
  journal={arXiv preprint arXiv:2504.02792},
  year={2025}
}

@article{pai2025mimic,
  title={mimic-video: Video-action models for generalizable robot control beyond vlas},
  author={Pai, Jonas and Achenbach, Liam and Montesinos, Victoriano and Forrai, Benedek and Mees, Oier and Nava, Elvis},
  journal={arXiv preprint arXiv:2512.15692},
  year={2025}
}

@article{yan2026s,
  title={S-vam: Shortcut video-action model by self-distilling geometric and semantic foresight},
  author={Yan, Haodong and Zhong, Zhide and Zhu, Jiaguan and He, Junjie and Yuan, Weilin and Song, Wenxuan and Gong, Xin and Cai, Yingjie and Zhao, Guanyi and Yan, Xu and others},
  journal={arXiv preprint arXiv:2603.16195},
  year={2026}
}

@article{ma2026dit4dit,
  title={Dit4dit: Jointly modeling video dynamics and actions for generalizable robot control},
  author={Ma, Teli and Zheng, Jia and Wang, Zifan and Jiang, Chunli and Cui, Andy and Liang, Junwei and Yang, Shuo},
  journal={arXiv preprint arXiv:2603.10448},
  year={2026}
}

@article{chen2026lawam,
  title={LaWAM: Latent World Action Models for Efficient Dynamics-Aware Robot Policies},
  author={Chen, Jialei and Wang, Kai and Chen, Kang and Chen, Shuaihang and Gao, Feng and Tang, Wenhao and Li, Zhiyuan and Liu, Weilin and Yao, Zhuyu and Li, Boxun and others},
  journal={arXiv preprint arXiv:2606.15768},
  year={2026}
}

@inproceedings{rectifiedflow,
  title={Scaling rectified flow transformers for high-resolution image synthesis},
  author={Esser, Patrick and Kulal, Sumith and Blattmann, Andreas and Entezari, Rahim and M{\"u}ller, Jonas and Saini, Harry and Levi, Yam and Lorenz, Dominik and Sauer, Axel and Boesel, Frederic and others},
  booktitle={Forty-first international conference on machine learning},
  year={2024}
}

@article{lipman2022flow,
  title={Flow matching for generative modeling},
  author={Lipman, Yaron and Chen, Ricky TQ and Ben-Hamu, Heli and Nickel, Maximilian and Le, Matt},
  journal={arXiv preprint arXiv:2210.02747},
  year={2022}
}

@article{xiang2024pandora,
  title={Pandora: Towards general world model with natural language actions and video states},
  author={Xiang, Jiannan and Liu, Guangyi and Gu, Yi and Gao, Qiyue and Ning, Yuting and Zha, Yuheng and Feng, Zeyu and Tao, Tianhua and Hao, Shibo and Shi, Yemin and others},
  journal={arXiv preprint arXiv:2406.09455},
  year={2024}
}

@article{kim2025freeaction,
  title={FreeAction: Training-Free Techniques for Enhanced Fidelity of Trajectory-to-Video Generation},
  author={Kim, Seungwook and Lee, Seunghyeon and Cho, Minsu},
  journal={arXiv preprint arXiv:2509.24241},
  year={2025}
}

@inproceedings{liu2026robotransfer,
  title={RoboTransfer: Controllable Geometry-Consistent Video Diffusion for Manipulation Policy Transfer},
  author={Liu, Liu and Wang, Xiaofeng and Zhao, Guosheng and Li, Keyu and Qin, Wenkang and Zhu, Jiagang and Qiu, Jiaxiong and Huang, Guan and Su, Zhizhong},
  booktitle={Proceedings of the IEEE/CVF Conference on Computer Vision and Pattern Recognition},
  pages={1410--1420},
  year={2026}
}

@article{alhaija2025cosmos,
  title={Cosmos-transfer1: Conditional world generation with adaptive multimodal control},
  author={Alhaija, Hassan Abu and Alvarez, Jose and Bala, Maciej and Cai, Tiffany and Cao, Tianshi and Cha, Liz and Chen, Joshua and Chen, Mike and Ferroni, Francesco and Fidler, Sanja and others},
  journal={arXiv preprint arXiv:2503.14492},
  year={2025}
}

@article{zhang2026qwen,
  title={Qwen-RobotWorld Technical Report: Unifying Embodied World Modeling through Language-Conditioned Video Generation},
  author={Zhang, Jie and Chen, Xiaoyue and Chen, Anzhe and Lv, Chenxu and Li, Deqing and Zhou, Gengze and Yin, Hang and Yuan, Haoqi and Li, Haoyang and Li, Jiahao and others},
  journal={arXiv preprint arXiv:2606.17030},
  year={2026}
}

@article{li2025mimicdreamer,
  title={Mimicdreamer: Aligning human and robot demonstrations for scalable vla training},
  author={Li, Haoyun and Zhang, Ivan and Ouyang, Runqi and Wang, Xiaofeng and Zhu, Zheng and Yang, Zhiqin and Zhang, Zhentao and Wang, Boyuan and Ni, Chaojun and Qin, Wenkang and others},
  journal={arXiv preprint arXiv:2509.22199},
  year={2025}
}

@article{ni2025recondreamer,
  title={Recondreamer-rl: Enhancing reinforcement learning via diffusion-based scene reconstruction},
  author={Ni, Chaojun and Zhao, Guosheng and Wang, Xiaofeng and Zhu, Zheng and Qin, Wenkang and Chen, Xinze and Jia, Guanghong and Huang, Guan and Mei, Wenjun},
  journal={arXiv preprint arXiv:2508.08170},
  year={2025}
}

@article{wang2025humandreamer,
  title={Humandreamer-x: Photorealistic single-image human avatars reconstruction via gaussian restoration},
  author={Wang, Boyuan and Ouyang, Runqi and Wang, Xiaofeng and Zhu, Zheng and Zhao, Guosheng and Ni, Chaojun and Zhang, Xiaopei and Huang, Guan and Ren, Yijie and Liu, Lihong and others},
  journal={arXiv preprint arXiv:2504.03536},
  year={2025}
}

@inproceedings{zhao2025drivedreamer4d,
  title={Drivedreamer4d: World models are effective data machines for 4d driving scene representation},
  author={Zhao, Guosheng and Ni, Chaojun and Wang, Xiaofeng and Zhu, Zheng and Zhang, Xueyang and Wang, Yida and Huang, Guan and Chen, Xinze and Wang, Boyuan and Zhang, Youyi and others},
  booktitle={Proceedings of the computer vision and pattern recognition conference},
  pages={12015--12026},
  year={2025}
}

@article{dong2025emma,
  title={Emma: Generalizing real-world robot manipulation via generative visual transfer},
  author={Dong, Zhehao and Wang, Xiaofeng and Zhu, Zheng and Wang, Yirui and Wang, Yang and Zhou, Yukun and Wang, Boyuan and Ni, Chaojun and Ouyang, Runqi and Qin, Wenkang and others},
  journal={arXiv preprint arXiv:2509.22407},
  year={2025}
}

@article{wang2025embodiedreamer,
  title={Embodiedreamer: Advancing real2sim2real transfer for policy training via embodied world modeling},
  author={Wang, Boyuan and Meng, Xinpan and Wang, Xiaofeng and Zhu, Zheng and Ye, Angen and Wang, Yang and Yang, Zhiqin and Ni, Chaojun and Huang, Guan and Wang, Xingang},
  journal={arXiv preprint arXiv:2507.05198},
  year={2025}
}

@article{jiang2026wovr,
  title={Wovr: World models as reliable simulators for post-training vla policies with rl},
  author={Jiang, Zhennan and Zhou, Shangqing and Jiang, Yutong and Huang, Zefang and Wei, Mingjie and Chen, Yuhui and Zhou, Tianxing and Guo, Zhen and Lin, Hao and Zhang, Quanlu and others},
  journal={arXiv preprint arXiv:2602.13977},
  year={2026}
}

@inproceedings{umt5,
  title={UniMax: Fairer and More Effective Language Sampling for Large-Scale Multilingual Pretraining},
  author={Chung, Hyung Won and Garcia, Xavier and Roberts, Adam and Tay, Yi and Firat, Orhan and Narang, Sharan and Constant, Noah},
  booktitle={The Eleventh International Conference on Learning Representations},
  year={2023}
}

@article{bu2025agibot,
  title={Agibot world colosseo: A large-scale manipulation platform for scalable and intelligent embodied systems},
  author={Bu, Qingwen and Cai, Jisong and Chen, Li and Cui, Xiuqi and Ding, Yan and Feng, Siyuan and Gao, Shenyuan and He, Xindong and Hu, Xuan and Huang, Xu and others},
  journal={arXiv preprint arXiv:2503.06669},
  year={2025}
}

@inproceedings{liu2025rdt,
  title={Rdt-1b: a diffusion foundation model for bimanual manipulation},
  author={Liu, Songming and Wu, Lingxuan and Li, Bangguo and Tan, Hengkai and Chen, Huayu and Wang, Zhengyi and Xu, Ke and Su, Hang and Zhu, Jun},
  booktitle={International Conference on Learning Representations},
  volume={2025},
  pages={29982--30009},
  year={2025}
}

@article{wu2024robomind,
  title={Robomind: Benchmark on multi-embodiment intelligence normative data for robot manipulation},
  author={Wu, Kun and Hou, Chengkai and Liu, Jiaming and Che, Zhengping and Ju, Xiaozhu and Yang, Zhuqin and Li, Meng and Zhao, Yinuo and Xu, Zhiyuan and Yang, Guang and others},
  journal={arXiv preprint arXiv:2412.13877},
  year={2024}
}

@article{tan2025anypos,
  title={Anypos: Automated task-agnostic actions for bimanual manipulation},
  author={Tan, Hengkai and Feng, Yao and Mao, Xinyi and Huang, Shuhe and Liu, Guodong and Hao, Zhongkai and Su, Hang and Zhu, Jun},
  journal={arXiv preprint arXiv:2507.12768},
  year={2025}
}

@inproceedings{Robodreamer,
  title={RoboDreamer: learning compositional world models for robot imagination},
  author={Zhou, Siyuan and Du, Yilun and Chen, Jiaben and Li, Yandong and Yeung, Dit-Yan and Gan, Chuang},
  booktitle={Proceedings of the 41st International Conference on Machine Learning},
  pages={61885--61896},
  year={2024}
}

@article{gigaworld1,
  title={GigaWorld-1: A Roadmap to Build World Models for Robot Policy Evaluation}, 
  author={Team, GigaWorld and Ma, Angyuan and Wang, Boyuan and Li, Bohan and Ni, Chaojun and Li, Guo  and Huang, Guan and Zhao, Guosheng and Li, Hao and Li, Hengtao and others},
  journal={arXiv preprint arXiv:2607.02642},
  year={2026}
}

@inproceedings{recondreamer++,
  title={Recondreamer++: Harmonizing generative and reconstructive models for driving scene representation},
  author={Zhao, Guosheng and Wang, Xiaofeng and Ni, Chaojun and Zhu, Zheng and Qin, Wenkang and Huang, Guan and Wang, Xingang},
  booktitle={Proceedings of the IEEE/CVF International Conference on Computer Vision},
  pages={26718--26728},
  year={2025}
}

@article{unidrivedreamer,
  title={UniDriveDreamer: A Single-Stage Multimodal World Model for Autonomous Driving},
  author={Zhao, Guosheng and Wang, Yaozeng and Wang, Xiaofeng and Zhu, Zheng and Yu, Tingdong and Huang, Guan and Zai, Yongchen and Jiao, Ji and Xue, Changliang and Wang, Xiaole and others},
  journal={arXiv preprint arXiv:2602.02002},
  year={2026}
}

@article{drivedreamerpolicy,
  title={Drivedreamer-policy: A geometry-grounded world-action model for unified generation and planning},
  author={Zhou, Yang and Wang, Xiaofeng and Shao, Hao and Wang, Letian and Zhao, Guosheng and Shao, Jiangnan and Zhu, Jiagang and Yu, Tingdong and Zhu, Zheng and Huang, Guan and others},
  journal={arXiv preprint arXiv:2604.01765},
  year={2026}
}

@article{wang2026egovid,
  title={EgoVid-5M: A Large-Scale Video-Action Dataset for Egocentric Videos Generation},
  author={Wang, Xiaofeng and Zhao, Kang and Liu, Feng and Wang, Jiayu and Zhao, Guosheng and Bao, Xiaoyi and Zhu, Zheng and Zhang, Yingya},
  journal={Advances in Neural Information Processing Systems},
  volume={38},
  year={2026}
}

@inproceedings{zhu2025aether,
  title={Aether: Geometric-aware unified world modeling},
  author={Zhu, Haoyi and Wang, Yifan and Zhou, Jianjun and Chang, Wenzheng and Zhou, Yang and Li, Zizun and Chen, Junyi and Shen, Chunhua and Pang, Jiangmiao and He, Tong},
  booktitle={Proceedings of the IEEE/CVF International Conference on Computer Vision},
  pages={8535--8546},
  year={2025}
}

@misc{karpathy2026autoresearch,
  author  = {Karpathy, Andrej},
  title   = {autoresearch},
  url     = {https://github.com/karpathy/autoresearch},
  year    = {2026}
}

@inproceedings{liu2025timestep,
  title={Timestep Embedding Tells: It's Time to Cache for Video Diffusion Model},
  author={Liu, Feng and Zhang, Shiwei and Wang, Xiaofeng and Wei, Yujie and Qiu, Haonan and Zhao, Yuzhong and Zhang, Yingya and Ye, Qixiang and Wan, Fang},
  booktitle={Proceedings of the Computer Vision and Pattern Recognition Conference},
  pages={7353--7363},
  year={2025}
}

@article{zhu2024sora,
  title={Is sora a world simulator? a comprehensive survey on general world models and beyond},
  author={Zhu, Zheng and Wang, Xiaofeng and Zhao, Wangbo and Min, Chen and Li, Bohan and Deng, Nianchen and Dou, Min and Wang, Yuqi and Shi, Botian and Wang, Kai and others},
  journal={arXiv preprint arXiv:2405.03520},
  year={2024}
}

@article{wang2024worlddreamer,
  title={Worlddreamer: Towards general world models for video generation via predicting masked tokens},
  author={Wang, Xiaofeng and Zhu, Zheng and Huang, Guan and Wang, Boyuan and Chen, Xinze and Lu, Jiwen},
  journal={arXiv preprint arXiv:2401.09985},
  year={2024}
}

@article{ye2025vla,
  title={Vla-r1: Enhancing reasoning in vision-language-action models},
  author={Ye, Angen and Zhang, Zeyu and Wang, Boyuan and Wang, Xiaofeng and Zhang, Dapeng and Zhu, Zheng},
  journal={arXiv preprint arXiv:2510.01623},
  year={2025}
}

@article{ye2026halo,
  title={HALO-WA: Hybrid-Attention Latent-Guided Online Reinforcement Learning for World-Action Models},
  author={Ye, Angen and Ke, Weijie and Wang, Xiaofeng and Chen, Xinze and Ni, Chaojun and Zhao, Guosheng and Wang, Boyuan and Zhu, Zheng and Xie, Junjie and Zhang, Dapeng},
  journal={arXiv preprint arXiv:2607.04265},
  year={2026}
}

@article{lv2026viva,
  title={ViVa: A Video-Generative Value Model for Robot Reinforcement Learning},
  author={Lv, Jindi and Li, Hao and Li, Jie and Kong, Fankun and Wang, Yang and Yi, Pengfei and Nie, Yifei and Wang, Xiaofeng and Zhu, Zheng and Ni, Chaojun and others},
  journal={arXiv preprint arXiv:2604.08168},
  year={2026}
}

@article{wang2026reconphys,
  title={ReconPhys: Reconstruct Appearance and Physical Attributes from Single Video},
  author={Wang, Boyuan and Wang, Xiaofeng and Li, Yongkang and Zhu, Zheng and Chang, Yifan and Ye, Angen and Zhao, Guosheng and Ni, Chaojun and Huang, Guan and Ren, Yijie and others},
  journal={arXiv preprint arXiv:2604.07882},
  year={2026}
}

\end{document}